\documentclass[10pt,twocolumn,letterpaper]{article}

\usepackage{cvpr}
\usepackage{times}
\usepackage{epsfig}
\usepackage{graphicx}
\usepackage{amsmath}
\usepackage{cases}
\usepackage{overpic}
\graphicspath{{figures/}}
\usepackage[sort,numbers]{natbib}
\usepackage{multirow}
\usepackage{tabularx}
\usepackage{subfigure}
\usepackage{rotating}
\usepackage{bm}
\usepackage{enumitem}
\usepackage{epstopdf}
\usepackage[pagebackref=true,breaklinks=true,letterpaper=true,colorlinks,bookmarks=false]{hyperref}

\cvprfinalcopy % *** Uncomment this line for the final submission

\def\cvprPaperID{1440} % *** Enter the CVPR Paper ID here

\ifcvprfinal\fi
\begin{document}

%%%%%%%%% TITLE
\title{Learning a Single Convolutional Super-Resolution Network for\\ Multiple Degradations}
\author{Kai Zhang$^{1,2,3}$,  Wangmeng Zuo$^{1}$, Lei Zhang$^2$\\
$^1$School of Computer Science and Technology, Harbin Institute of Technology, Harbin, China\\
$^2$Dept. of Computing, The Hong Kong Polytechnic University, Hong Kong, China\\
$^3$DAMO Academy, Alibaba Group\\
{\tt\small cskaizhang@gmail.com, wmzuo@hit.edu.cn, cslzhang@comp.polyu.edu.hk}
}
\maketitle
\thispagestyle{empty}
%%%%%%%%% ABSTRACT
\begin{abstract}
Recent years have witnessed the unprecedented success of deep convolutional neural networks (CNNs) in single image super-resolution (SISR).
However, existing CNN-based SISR methods mostly assume that a low-resolution (LR) image is bicubicly downsampled from a high-resolution (HR) image,
thus inevitably giving rise to poor performance when the true degradation does not follow this assumption.
Moreover, they lack scalability in learning a single model to non-blindly deal with multiple degradations. To address these issues, we propose a general framework with dimensionality stretching strategy that enables a single convolutional super-resolution network to take two key factors of the SISR degradation process, i.e., blur kernel and noise level, as input. Consequently, the super-resolver can handle multiple and even spatially variant degradations, which significantly improves the practicability. Extensive experimental results on synthetic and real LR images show that the proposed convolutional super-resolution network not only can produce favorable results on multiple degradations but also is computationally efficient, providing a highly effective and scalable solution to practical SISR applications.

\end{abstract}

\vspace{-0.2cm}
\section{Introduction}
Single image super-resolution (SISR) aims to recover a high-resolution (HR) version of a low-resolution (LR) input.
As a classical problem, SISR is still an active yet challenging research topic in the field of computer vision due to its ill-poseness nature and high practical values~\cite{baker2002limits}.
In the typical SISR framework, an LR image $\textbf{\emph{y}}$ is modeled as the output of the following degradation process:
\begin{equation}\label{eq1}
  \textbf{\emph{y}}=(\textbf{\emph{x}} \otimes \textbf{\emph{k}})\downarrow_s + ~ \textbf{\emph{n}},
\end{equation}
where $\textbf{\emph{x}} \otimes \textbf{\emph{k}}$ represents the convolution between a blur kernel $\textbf{\emph{k}}$ and a latent HR image $\textbf{\emph{x}}$,
$\downarrow_s$ is a subsequent downsampling operation with scale factor $s$, and $\boldsymbol{n}$ usually is additive white Gaussian noise (AWGN) with standard deviation (noise level) $\sigma$.

SISR methods can be broadly classified into three categories, \ie, interpolation-based methods, model-based optimization methods and discriminative learning methods. Interpolation-based methods such as nearest-neighbor, bilinear and bicubic interpolators are simple and efficient but have very limited performance.
By exploiting powerful image priors (\eg, the non-local self-similarity prior~\cite{mairal2009non,dong2013nonlocally}, sparsity prior~\cite{yang2010image} and denoiser prior~\cite{egiazarian2015single,zhang2017learning,bigdeli2017deep}), model-based optimization methods
are flexible to reconstruct relative high-quality HR images, but they usually involve a time-consuming optimization procedure. Although the integration of convolutional neural network (CNN) denoiser prior and model-based optimization can improve the efficiency to some extent,
it still suffers from the typical drawbacks of model-based optimization methods, \eg, it is not in an end-to-end learning manner and involves hand-designed parameters~\cite{zhang2017learning}.
As an alternative, discriminative learning methods have attracted considerable attentions due to
their favorable SISR performance in terms of effectiveness and efficiency.
Notably, recent years have witnessed a dramatic upsurge of using CNN for SISR.

In this paper, we focus on discriminative CNN methods for SISR so as to exploit the merits of CNN, such as the fast speed by parallel computing, high accuracy by end-to-end training, and tremendous advances in training and designing networks~\cite{lecun2015deep,jaderberg2015spatial,goodfellow2014generative,he2016deep}.
While several SISR models based on discriminative CNN have reported impressive results, they suffer from a common drawback: their models are specialized for a single simplified degradation (\eg, bicubic degradation) and lack scalability to handle multiple degradations by using a single model. Because the practical degradation of SISR is much more complex~\cite{romano2017raisr,yang2014single}, the performance of learned CNN models may deteriorate seriously when the assumed degradation deviates from the true one, making them less effective in practical scenarios. It has been pointed out that the blur kernel plays a vital role for the success of SISR methods and the mismatch of blur kernels will largely deteriorate the final SISR results~\cite{efrat2013accurate}.
However, little work has been done on how to design a CNN to address this crucial issue.

Given the facts above,
it is natural to raise the following questions, which are the focus of our paper:
(i) Can we learn a single model to effectively handle multiple and even spatially variant degradations?
(ii) Is it possible to use synthetic data to train a model with high practicability?
This work aims to make one of the first attempts towards answering these two questions.

To answer the first question, we revisit and analyze the general model-based SISR methods under the maximum a posteriori (MAP) framework. Then we argue that one may tackle this issue by taking LR input, blur kernel and noise level as input to CNN but their dimensionality mismatch makes it difficult to design a single convolutional super-resolution network. In view of this, we introduce a dimensionality stretching strategy which facilitates the network to handle multiple and even spatially variant degradations with respect to blur kernel and noise. To the best of our knowledge, there is no attempt to consider  both the blur kernel and noise for SISR via training a single CNN model.

For the second question, we will show that it is possible to learn a practical super-resolver using synthetic data. To this end, a large variety of degradations with different combinations of blur kernels and noise levels are sampled to cover the degradation space. In a practical scenario, even the degradation is more complex (\eg, the noise is non-AWGN), we can select the best fitted degradation model rather than the bicubic degradation to produce a better result. It turns out that, by choosing a proper degradation, the learned SISR model can yield perceptually convincing results on real LR images. It should be noted that we make no effort to use specialized network architectures but use the plain CNN as in~\cite{dong2016image,shi2016real}.% since our solution is a general framework.

The main contributions of this paper are summarized in the following:
\begin{itemize}
  \item We propose a simple yet effective and scalable deep CNN framework for SISR. The proposed model goes beyond the widely-used bicubic degradation assumption and works for multiple and even spatially variant degradations, thus making a substantial step towards developing a practical CNN-based super-resolver for real applications.

  \item  We propose a novel dimensionality stretching strategy to address the dimensionality mismatch between LR input image, blur kernel and noise level. Although this strategy is proposed for SISR, it is general and can be extended to other tasks such as deblurring.

  \item We show that the proposed convolutional super-resolution network learned from synthetic training data can not only produce competitive results against state-of-the-art SISR methods on synthetic LR images but also give rise to visually plausible results on real LR images.

\end{itemize}

\section{Related Work}
%In this section, we give a brief review of existing CNN-based SISR methods.

The first work of using CNN to solve SISR can be traced back to~\cite{dong2014learning} where a three-layer super-resolution network (SRCNN) was proposed.
In the extended work~\cite{dong2016image}, the authors investigated the impact of depth on super-resolution and empirically showed that the difficulty
of training deeper model hinders the performance improvement of CNN super-resolvers.
To overcome the training difficulty, Kim~\etal~\cite{kim2015accurate} proposed a very deep super-resolution (VDSR)
method with residual learning strategy. Interestingly, they showed that VDSR can handle multiple scales super-resolution.
By analyzing the relation between CNN and MAP inference, Zhang~\etal~\cite{zhang2017beyond} pointed out that CNN mainly model the prior information and they empirically demonstrated that a single model can handle multiple scales super-resolution, image deblocking and image denoising.
While achieving good performance, the above methods take the bicubicly interpolated LR image as input, which not only suffers from high computational cost but also hinders the effective expansion of receptive field.

To improve the efficiency, some researchers resort to directly manipulating the LR input and adopting an upscaling operation at the end of the network.
Shi~\etal~\cite{shi2016real} introduced an efficient sub-pixel convolution layer to upscale the LR feature maps into HR images.
Dong~\etal~\cite{dong2016accelerating} adopted a deconvolution layer at the end of the network to perform upsampling.
Lai~\etal~\cite{lai2017deep} proposed a Laplacian pyramid super-resolution network (LapSRN) that takes an LR image as input and progressively predicts
the sub-band residuals with transposed convolutions in a coarse-to-fine manner.
To improve the perceptual quality at a large scale factor, Ledig~\etal~\cite{ledig2016photo} proposed a generative adversarial network~\cite{goodfellow2014generative} based super-resolution (SRGAN) method.
In the generator network of SRGAN, two sub-pixel convolution layers are used to efficiently upscale the LR input by a factor of 4.

Although various techniques have been proposed for SISR, the above CNN-based methods are tailored to the widely-used settings of bicubic degradation, neglecting their limited applicability for practical scenarios.
An interesting line of CNN-based methods which can go beyond bicubic degradation adopt a CNN denoiser to solve SISR via model-based optimization framework~\cite{zhang2017learning,bigdeli2017deep,meinhardt2017learning}.
For example, the method proposed in~\cite{zhang2017learning} can handle the widely-used Gaussian degradation as in~\cite{dong2013nonlocally}.
However, manually selecting the hyper-parameters for different degradations is not a trivial task~\cite{romano2017little}.
As a result, it is desirable to learn a single SISR model which can handle multiple degradations with high practicability. This paper attempts to give a positive answer.

Due to the limited space, we can only discuss some of the related works here. Other CNN-based SISR methods can be found in~\cite{ren2015shepard,kim2016deeply,yang2017deep,johnson2016perceptual,chen2015learning,tai2017memnet,lim2017enhanced,Timofte2017CVPR,shi2017structure,taiimage,zhang2018residual}.

\section{Method}

\subsection{Degradation Model}

Before solving the problem of SISR, it is important to have a clear understanding of the degradation model which is not limited to Eqn.~\eqref{eq1}.
Another practical degradation model can be given by
\begin{equation}\label{eq_degradation2}
  \textbf{\emph{y}}=(\textbf{\emph{x}}\downarrow_s) \otimes \textbf{\emph{k}} + ~ \textbf{\emph{n}}.
\end{equation}
When $\downarrow$ is the bicubic downsampler,
Eqn.~\eqref{eq_degradation2} corresponds to a deblurring problem followed by a SISR problem with bicubic degradation.
Thus, it can benefit from existing deblurring methods and bicubic degradation based SISR methods.
Due to limited space, we only consider the more widely assumed degradation model given in Eqn.~\eqref{eq1}.
Nevertheless, our method is general and can be easily extended to handle Eqn.~\eqref{eq_degradation2}.
In the following, we make a short discussion on blur kernel $\textbf{\emph{k}}$, noise $\textbf{\emph{n}}$ and downsampler $\downarrow$.

\vspace{-0.35cm}
\paragraph{Blur kernel.} Different from image deblurring, the blur kernel setting of SISR is usually simple. The most popular choice is isotropic Gaussian blur kernel parameterized by standard deviation or kernel width~\cite{dong2013nonlocally,yang2014single}.
In~\cite{riegler2015conditioned}, anisotropic Gaussian blur kernels are also used.
In practice, more complex blur kernel models used in deblurring task, such as motion blur~\cite{Boracchi2012}, can be further considered.
Empirical and theoretical analyses have revealed that the influence of an accurate blur kernel is much larger than that of sophisticated image priors~\cite{efrat2013accurate}.
Specifically, when the assumed kernel is smoother than the true kernel, the recovered image is over-smoothed. Most of SISR methods actually favor for such case. On the other hand, when the assumed kernel is sharper than the true kernel, high frequency ringing artifacts will appear.

\vspace{-0.35cm}
\paragraph{Noise.} While being of low-resolution, the LR images are usually also noisy. Directly super-resolving the noisy input without noise removal would amplify the unwanted noise, resulting in visually unpleasant results. To address this problem, the straightforward way is to perform denoising first and then enhance the resolution. However, the denoising pre-processing step tends to lose detail information and would deteriorate the subsequent super-resolution performance~\cite{singh2014super}.
Thus, it would be highly desirable to jointly perform denoising and super-resolution.

\vspace{-0.35cm}
\paragraph{Downsampler.} Existing literatures have considered two types of downsamplers, including direct downsampler~\cite{he2011single,peleg2014statistical,yang2014single,dong2013nonlocally,zhang2015revisiting} and bicubic downsampler~\cite{glasner2009super,yang2010image,cui2014deep,timofte2014a+,freeman2011markov,efrat2013accurate}.
In this paper, we consider the bicubic downsampler since when $\textbf{\emph{k}}$ is delta kernel and the noise level is zero, Eqn.~\eqref{eq1} turns into the widely-used bicubic degradation model.
It should be pointed out that, different from blur kernel and noise which vary in a general degradation model, downsampler is assumed to be fixed.

\begin{figure}[!tbp]
\begin{center}
\begin{overpic}[width=0.48\textwidth]{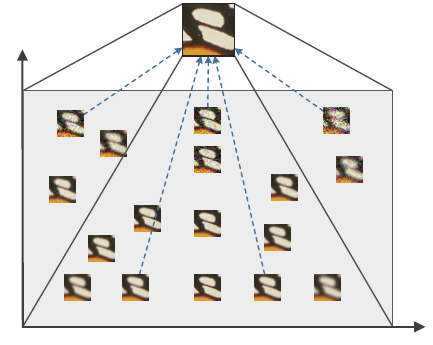}
\put(40,0.5){\color{black}{\small \textbf{Blur Kernel}}}
\begin{turn}{90}
%\begin{sideways}
\put(32,-4.2){\color{black}{\small \textbf{Noise}}}
%\end{sideways}
\end{turn}
\end{overpic}
\caption{An illustration of different degradations for SISR. The scale factor is 2. The general degradation models of Eqns.~\eqref{eq1} and~\eqref{eq_degradation2} assume an HR image actually can degrade into many LR images, whereas bicubic degradation model assumes an HR image corresponds to a single LR image.}\label{fig_dm1}
\end{center}\vspace{-0.65cm}
\end{figure}

Though blur kernel and noise have been recognized as key factors for the success of SISR and several methods have been proposed to consider those two factors, there has been little effort towards simultaneously considering blur kernel and noise in a single CNN framework. It is a challenging task since the degradation space with respect to blur kernel and noise is rather large (see Figure~\ref{fig_dm1} as an example).
One relevant work is done by Zhang~\etal~\cite{zhang2017learning}; nonetheless, their method is essentially a model-based optimization method and thus suffers from several drawbacks as mentioned previously.
In another related work, Riegler~\etal~\cite{riegler2015conditioned} exploited the blur kernel information into the SISR model.
Our method differs from~\cite{riegler2015conditioned} on two major aspects. First, our method considers a more general degradation model. Second, our method exploits a more effective way to parameterize the degradation model.

\subsection{A Perspective from MAP Framework}

Though existing CNN-based SISR methods are not necessarily derived under the traditional MAP framework, they have the same goal. We revisit and analyze the general MAP framework of SISR, aiming to find the intrinsic connections between the MAP principle and the working mechanism of CNN. Consequently, more insights on CNN architecture design can be obtained.

Due to the ill-posed nature of SISR, regularization needs to be imposed to constrain the solution.
Mathematically, the HR counterpart of an LR image $\textbf{\emph{y}}$ can be estimated by solving the following MAP problem
\begin{equation}\label{eq33}
 \boldsymbol{\hat{\textbf{\emph{x}}}} = \mathop{\arg\min}\displaystyle_{\textbf{\emph{x}}}\frac{1}{2\sigma^2}\|(\textbf{\emph{x}} \otimes \textbf{\emph{k}})\downarrow_s-\textbf{\emph{y}}\|^2  + \lambda \Phi(\textbf{\emph{x}})
\end{equation}
where $\frac{1}{2\sigma^2}\|(\textbf{\emph{x}} \otimes \textbf{\emph{k}})\downarrow_s-\textbf{\emph{y}}\|^2$ is the data fidelity term, $\Phi(\textbf{\emph{x}})$ is the
regularization term (or prior term) and $\lambda$ is the trade-off parameter.
Simply speaking, Eqn.~\eqref{eq33} conveys two points: (i) the estimated solution should not only accord with the degradation process but also have the desired property of clean HR images; (ii)
$\boldsymbol{\hat{\textbf{\emph{x}}}}$ is a function of LR image $\textbf{\emph{y}}$, blur kernel $\textbf{\emph{k}}$, noise level $\sigma$, and trade-off parameter $\lambda$.
Therefore, the MAP solution of (non-blind) SISR can be formulated as
\begin{equation}\label{eq4}
 \boldsymbol{\hat{\textbf{\emph{x}}}} = \mathcal{F}(\textbf{\emph{y}}, \textbf{\emph{k}},\sigma, \lambda; \Theta)
\end{equation}
where $\Theta$ denotes the parameters of the MAP inference.

By treating CNN as a discriminative learning solution to Eqn.~\eqref{eq4}, we can have the following insights.
\begin{itemize}[fullwidth]
  \item Because the data fidelity term corresponds to the degradation process, accurate modeling of the degradation plays a key role for the success of SISR. However, existing CNN-based SISR methods with bicubic degradation actually aim to solve the following problem
  \begin{equation}\label{eq5}
 \boldsymbol{\hat{\textbf{\emph{x}}}} = \mathop{\arg\min}\displaystyle_{\textbf{\emph{x}}}\|\textbf{\emph{x}}\downarrow_s-\textbf{\emph{y}}\|^2  + \Phi(\textbf{\emph{x}}).
\end{equation}
Inevitably, their practicability is very limited.

\item To design a more practical SISR model, it is preferable to learn a mapping function like Eqn.~\eqref{eq4}, which covers more extensive degradations. It should be stressed that, since $\lambda$ can be absorbed into $\sigma$, Eqn.~\eqref{eq4} can be reformulated as
\begin{equation}\label{eq6}
 \boldsymbol{\hat{\textbf{\emph{x}}}} = \mathcal{F}(\textbf{\emph{y}}, \textbf{\emph{k}},\sigma; \Theta).
\end{equation}

\item Considering that the MAP framework (Eqn.~\eqref{eq33}) can perform generic image super-resolution with the same image prior, it is intuitive to jointly perform denoising and SISR in a unified CNN framework. Moreover, the work~\cite{zhang2017beyond} indicates that the parameters of the MAP inference mainly model the prior; therefore, CNN has the capacity to deal with multiple degradations via a single model.
\end{itemize}

From the viewpoint of MAP framework, one can see that the goal of SISR is to learn a mapping function
$\boldsymbol{\hat{\textbf{\emph{x}}}} = \mathcal{F}(\textbf{\emph{y}}, \textbf{\emph{k}},\sigma; \Theta)$ rather than $\boldsymbol{\hat{\textbf{\emph{x}}}} = \mathcal{F}(\textbf{\emph{y}}; \Theta)$.
However, it is not an easy task to directly model $\boldsymbol{\hat{\textbf{\emph{x}}}} = \mathcal{F}(\textbf{\emph{y}}, \textbf{\emph{k}},\sigma; \Theta)$ via CNN.
The reason lies in the fact that the three inputs $\textbf{\emph{y}}$, $\textbf{\emph{k}}$ and $\sigma$ have different dimensions.
In the next subsection, we will propose a simple dimensionality stretching strategy to resolve this problem.

\subsection{Dimensionality Stretching}
\begin{figure}[!tbp]
\begin{center}
%{\includegraphics[width=0.48\textwidth]{stretch102.pdf}}
\begin{overpic}[width=0.48\textwidth]{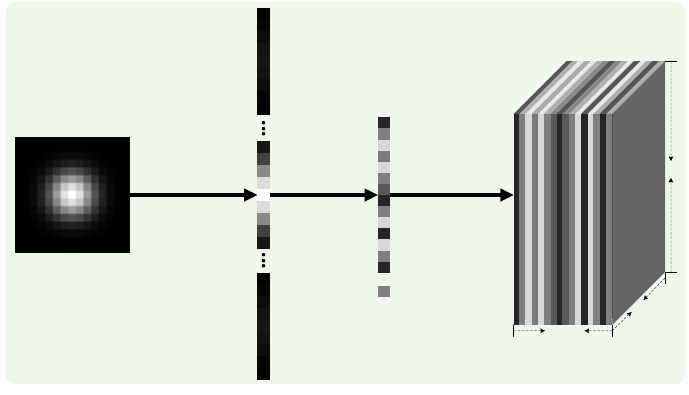}
\put(57.5,29.8){\color{black}{\scriptsize \textbf{Stretching}}}
\put(18.8,29.8){\color{black}{\scriptsize \textbf{Vectorization}}}
\put(3,18){\color{black}{\scriptsize \emph{\textbf{Blur Kernel}}}}
\put(43,29.8){\color{black}{\scriptsize \textbf{PCA}}}
\put(48.5,11){\color{black}{\scriptsize \emph{\textbf{Noise Level}}}}
\put(79.8,8.4){\color{black}{\tiny $t$$+$$1$}}
\put(74,4.8){\color{black}{\scriptsize \emph{\textbf{Degradation Maps}}}}
\put(91.7,11.8){\color{black}{\tiny $W$}}
\put(96.3,31.8){\color{black}{\tiny $H$}}
\end{overpic}
\caption{Schematic illustration of the dimensionality stretching strategy. For an LR image of size $W\times H$, the vectorized blur kernel is first projected onto a space of dimension $t$ and then stretched into a tensor $\mathcal{M}$ of size $W\times H\times (t+1)$ with the noise level.}\label{fig_na10}
\end{center}\vspace{-0.6cm}
\end{figure}

The proposed dimensionality stretching strategy is schematically illustrated in Figure~\ref{fig_na10}.
Suppose the inputs consist of a blur kernel of size $p$$\times$$p$, a noise level $\sigma$
and an LR image of size $W \times H \times C$, where $C$ denotes the number of channels. The blur kernel is first vectorized into a vector of size $p^2\times 1$ and then projected onto $t$-dimensional linear space by the PCA (Principal Component Analysis) technique.
After that, the concatenated low dimensional vector and the noise level, denoted by $\emph{\textbf{v}}$, are stretched into degradation maps $\mathcal{M}$ of size $W\times H\times (t+1)$, where all the elements of $i$-th map are $\emph{\textbf{v}}_i$.
By doing so, the degradation maps then can be concatenated with the LR image, making CNN possible to handle the three inputs.
Such a simple strategy can be easily exploited to deal with spatially variant degradations by considering the fact that the degradation maps can be non-uniform.

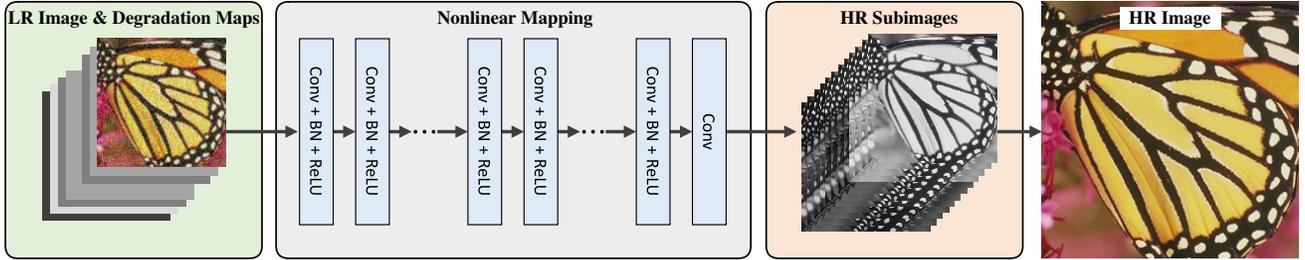
\begin{figure*}[!tbp]
\begin{center}
\begin{overpic}[width=0.99\textwidth]{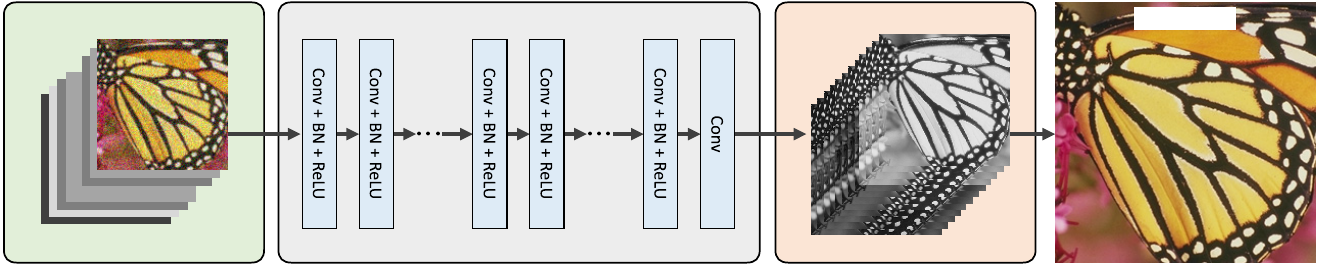}

\put(0.5,18.3){\color{black}{\scriptsize \textbf{LR Image \& Degradation Maps}}}
\put(33.5,18.3){\color{black}{\scriptsize \textbf{Nonlinear Mapping}}}
\put(64.5,18.3){\color{black}{\scriptsize \textbf{HR Subimages}}}
\put(86.7,18.3){\color{black}{\scriptsize \textbf{HR Image}}}
\end{overpic}
\caption{The architecture of the proposed convolutional super-resolution network. In contrast to other CNN-based SISR methods which only take the LR image as input and lack scalability to handle other degradations, the proposed network takes the concatenated LR image and degradation maps as input, thus allowing a single model to manipulate multiple and even spatially variant degradations.}\label{fig_na2}
\end{center}\vspace{-0.6cm}
\end{figure*}

\subsection{Proposed Network}

The proposed super-resolution network for multiple degradations, denoted by SRMD, is illustrated in Figure~\ref{fig_na2}.
As one can see, the distinctive feature of SRMD is that it takes the concatenated LR image and degradation maps as input.
To show the effectiveness of the dimensionality stretching strategy, we resort to plain CNN without complex architectural engineering.
Typically, to super-resolve an LR image with a scale factor of $s$, SRMD first takes the concatenated LR image and degradation maps of size $W \times H \times (C+t+1)$ as input. Then, similar to~\cite{kim2015accurate}, a cascade of $3\times 3$ convolutional layers are applied to perform the non-linear mapping.
Each layer is composed of three types of operations, including Convolution (Conv), Rectified Linear Units (ReLU)~\cite{krizhevsky2012imagenet}, and Batch Normalization (BN)~\cite{ioffe2015batch}.
Specifically, ``Conv + BN + ReLU'' is adopted for each convolutional layer except the last convolutional layer which consists of a single ``Conv'' operation.
Finally, a sub-pixel convolution layer~\cite{shi2016real} is followed by the last convolutional layer to convert multiple HR subimages of size $W \times H \times s^2C$ to a single HR image of size $sW \times sH \times C$.

For all scale factors 2, 3 and 4, the number of convolutional layers is set to $12$, and the number of feature maps in each layer is set to $128$.
We separately learn models for each scale factor.
In particular, we also learn the models for noise-free degradation, namely SRMDNF, by removing the connection of the noise level map in the first convolutional filter and fine-tuning with new training data.

It is worth pointing out that neither residual learning nor bicubicly interpolated LR image is used for the network design due to the following reasons.
First, with a moderate network depth and advanced CNN training and design such as ReLU~\cite{krizhevsky2012imagenet}, BN~\cite{ioffe2015batch} and Adam~\cite{kingma2014adam}, it is easy to train the network without the residual learning strategy.
Second, since the degradation involves noise, bicubicly interpolated LR image would aggravate the complexity of noise which in turn will increase the difficulty of training.

\subsection{Why not Learn a Blind Model?}
To enhance the practicability of CNN for SISR, it seems the most straightforward way is to learn a blind model with synthesized
training data by different degradations. However, such blind model does not perform as well as expected. % especially when the degradations are complex.
First, the performance deteriorates seriously when the blur kernel model is complex, \eg, motion blur.
This phenomenon can be explained by the following example. Given an HR image, a blur kernel and corresponding LR image, shifting the HR image to left by one pixel and shifting the blur kernel to right by one pixel would result in the same LR image. Thus, an LR image may correspond to different HR images with pixel shift. This in turn would aggravate the pixel-wise average problem~\cite{ledig2016photo}, typically leading to over-smoothed results.
Second, the blind model without specially designed architecture design has inferior generalization ability and performs poorly in real applications.

In contrast, non-blind model for multiple degradations suffers little from the pixel-wise average problem and has better generalization ability.
First, the degradation maps contain the warping information and thus can enable the network to have spatial transformation capability.
For clarity, one can treat the degradation maps induced by blur kernel and noise level as the output of a spatial transformer as in~\cite{jaderberg2015spatial}.
Second, by anchoring the model with degradation maps, the non-blind model generalizes easily to unseen degradations and has the ability to control the tradeoff between data fidelity term and regularization term.

\section{Experiments}

\subsection{Training Data Synthesis and Network Training}
Before synthesizing LR images according to Eqn.~\eqref{eq1}, it is necessary to define the blur kernels and noise level range, as well as providing a large-scale clean HR image set.

For the blur kernels,
we follow the kernel model of isotropic Gaussian with a fixed kernel width which has been proved practically feasible in SISR applications.
Specifically, the kernel width ranges are set to $[0.2, 2]$, $[0.2, 3]$ and $[0.2, 4]$ for scale factors 2, 3 and 4, respectively.
We sample the kernel width by a stride of $0.1$. The kernel size is fixed to 15$\times$15.
To further expand the degradation space, we also consider a more general kernel assumption, \ie, anisotropic Gaussian, which is characterized by a
Gaussian probability density function $\mathcal{N}(\mathbf{0}, \mathbf{\Sigma})$ with zero mean and varying covariance matrix $\mathbf{\Sigma}$~\cite{riegler2015conditioned}.
The space of such Gaussian kernel is determined by rotation angle of the eigenvectors of $\mathbf{\Sigma}$ and scaling of corresponding eigenvalues.
We set the rotation angle range to $[0, \pi]$. For the scaling of eigenvalues, it is set from $0.5$ to $6$, $8$ and $10$ for scale factors 2, 3 and 4, respectively.

Although we adopt the bicubic downsampler throughout the paper, it is straightforward to train a model with direct downsampler.
Alternatively, we can also include the degradations with direct downsampler by approximating it.
Specifically, given a blur kernel $\textbf{\emph{k}}_d$ under direct downsampler $\downarrow^{d}$, we can find the corresponding
blur kernel $\textbf{\emph{k}}_b$ under bicubic downsampler $\downarrow^{b}$ by solving the following problem with a data-driven method
\begin{equation}\label{eq_td1}
  \textbf{\emph{k}}_b = \mathop{\arg\min}\displaystyle_{\textbf{\emph{k}}_b}\|(\textbf{\emph{x}} \otimes \textbf{\emph{k}}_b)\downarrow^{b}_s - (\textbf{\emph{x}} \otimes \textbf{\emph{k}}_d)\downarrow^{d}_s \|^2, \quad  \forall \, \textbf{\emph{x}}.
\end{equation}
In this paper, we also include such degradations for scale factor 3.

\begin{figure}[!t]\vspace{0.2cm}
  \begin{center}
  \includegraphics[width=0.463\textwidth]{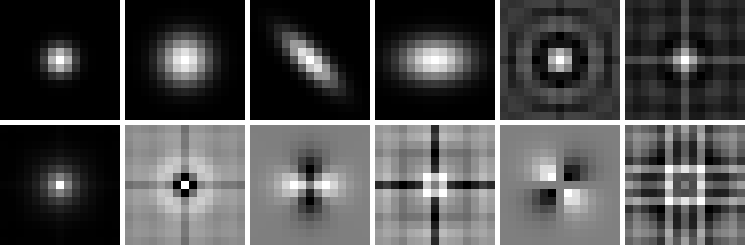}\\
  \caption{Visualization of six typical blur kernels (\textbf{fist row}) of isotropic Gaussian (first two), anisotropic Gaussian (middle two) and estimated ones for direct downsampler (last two) for scale factor 3 and PCA eigenvectors (\textbf{second row}) for the first six largest eigenvalues.}\label{fig_kernel}
  \end{center}\vspace{-0.6cm}
\end{figure}

Once the blur kernels are well-defined or learned, we then uniformly sample substantial kernels and aggregate them to learn the PCA projection matrix.
By preserving about $99.8\%$ of the energy, the kernels are projected onto a space of dimension $15$ (\ie, $t = 15$).
The visualization of some typical blur kernels for scale factor 3 and some PCA eigenvectors is shown in Figure~\ref{fig_kernel}.

For the noise level range, we set it as $[0, 75]$.
Because the proposed method operates on RGB channels rather than Y channel in YCbCr color space, we collect a large-scale color images for training,
including $400$ BSD~\cite{MartinFTM01} images, $800$ training images from DIV2K dataset~\cite{agustsson2017ntire} and $4,744$ images from WED dataset~\cite{ma2016gmad}.

Then, given an HR image, we synthesize LR image by blurring it with a blur kernel $\emph{\textbf{k}}$ and bicubicly downsampling it with a scale factor $s$, followed by an addition of AWGN with noise level $\sigma$.
The LR patch size is set to $40\times 40$ which means the corresponding HR patch sizes for scale factors 2, 3, and 4 are $80\times 80$, $120\times 120$ and $160\times 160$, respectively.

In the training phase, we randomly select a blur kernel and a noise level to synthesize an LR image and crop $N=128\times 3,000$ LR/HR patch pairs (along with the degradation maps) for each epoch.
We optimize the following loss function using Adam~\cite{kingma2014adam}
\begin{equation}\label{eq:loss}
  \mathcal{L}(\Theta) = \frac{1}{2N}\sum_{i=1}^N\|\mathcal{F}(\mathbf{y}_i, \mathbf{\mathcal{M}}_{i}; \Theta)  - \mathbf{x}_i \|^2.
\end{equation}
The mini-batch size is set to $128$. The learning rate starts from $10^{-3}$ and reduces to $10^{-4}$ when the training error stops decreasing. When the training error keeps unchanged in five sequential epochs, we merge the parameters of each batch normalization into the adjacent convolution filters. Then, a small learning rate of $10^{-5}$ is used for additional $100$ epochs to fine-tune the model. Since SRMDNF is obtained by fine-tuning SRMD, its learning rate is fixed to $10^{-5}$ for $200$ epochs.

We train the models in Matlab (R2015b) environment with MatConvNet package~\cite{vedaldi2015matconvnet} and an Nvidia Titan X Pascal GPU.
The training of a single SRMD model can be done in about two days.
The source code can be downloaded at \url{https://github.com/cszn/SRMD}.

\begin{table*}[!htbp]\scriptsize
\caption{Average PSNR and SSIM results for bicubic degradation on datasets Set5~\cite{bevilacqua2012low}, Set14~\cite{zeyde2010single}, BSD100~\cite{MartinFTM01} and Urban100~\cite{huang2015single}. The best two results are highlighted in \textcolor[rgb]{1.00,0.00,0.00}{red} and \textcolor[rgb]{0.00,0.00,1.00}{blue} colors, respectively.} \vspace{-0.16cm}%Red color indicates the best performance and blue color indicates the second best performance
\center
\begin{tabular}{|c|p{0.9cm}<{\centering}|p{1.4cm}<{\centering}|p{1.4cm}<{\centering}|p{1.4cm}<{\centering}|p{1.4cm}<{\centering}|p{1.4cm}<{\centering}|p{1.4cm}<{\centering}|p{1.4cm}<{\centering}|p{1.4cm}<{\centering}|}
  \hline
  % after \\: \hline or \cline{col1-col2} \cline{col3-col4} ...\multirow{4}{2cm}{This is a demo table}
 \multirow{2}{*}{Dataset} & Scale & Bicubic & SRCNN~\cite{dong2016image} & VDSR~\cite{kim2015accurate} & SRResNet~\cite{ledig2016photo}  & DRRN~\cite{taiimage}  & LapSRN~\cite{lai2017deep} & SRMD  & SRMDNF\\\cline{3-10}
  & Factor & \multicolumn{8}{c|}{PSNR / SSIM}\\ \hline\hline
  % &  & Bicubic & A$+$ & SelfEx & SRCNN  & VDSR  & JSCNN (Ours) \\ \hline
  &$\times 2$ & 33.64 / 0.929 &  36.62 / 0.953 &37.56 / 0.959 & --  & \textcolor[rgb]{0.00,0.00,1.00}{37.66} / \textcolor[rgb]{0.00,0.00,1.00}{0.959}  & 37.52 / 0.959 & 37.53 / 0.959  &  \textcolor[rgb]{1.00,0.00,0.00}{37.79} / \textcolor[rgb]{1.00,0.00,0.00}{0.960} \\
 Set5 &$\times 3$ & 30.39 / 0.868 &  32.74 / 0.908 &33.67 / 0.922 & --  & \textcolor[rgb]{0.00,0.00,1.00}{33.93} / \textcolor[rgb]{0.00,0.00,1.00}{0.923} & 33.82 / 0.922 & 33.86 / 0.923 & \textcolor[rgb]{1.00,0.00,0.00}{34.12} / \textcolor[rgb]{1.00,0.00,0.00}{0.925}\\
  &$\times 4$ & 28.42 / 0.810 &  30.48 / 0.863 &31.35 / 0.885 & \textcolor[rgb]{1.00,0.00,0.00}{32.05} / \textcolor[rgb]{0.00,0.00,1.00}{0.891}  & 31.58 / 0.886 & 31.54 / 0.885 & 31.59 / 0.887 & \textcolor[rgb]{0.00,0.00,1.00}{31.96} / \textcolor[rgb]{1.00,0.00,0.00}{0.893} \\\hline

    &$\times 2$ & 30.22 / 0.868 & 32.42 / 0.906&33.02 / 0.913 & --  & \textcolor[rgb]{0.00,0.00,1.00}{33.19} / 0.913  &  33.08 / 0.913 & 33.12 / \textcolor[rgb]{0.00,0.00,1.00}{0.914}  &  \textcolor[rgb]{1.00,0.00,0.00}{33.32} / \textcolor[rgb]{1.00,0.00,0.00}{0.915} \\
 Set14 &$\times 3$ & 27.53 / 0.774 & 29.27 / 0.821 &29.77 / 0.832 & --  & \textcolor[rgb]{0.00,0.00,1.00}{29.94} / \textcolor[rgb]{0.00,0.00,1.00}{0.834} & 29.89 / 0.834 & 29.84 / 0.833 & \textcolor[rgb]{1.00,0.00,0.00}{30.04} / \textcolor[rgb]{1.00,0.00,0.00}{0.837}\\
  &$\times 4$ & 25.99 / 0.702 &  27.48 / 0.751 &27.99 / 0.766 & \textcolor[rgb]{1.00,0.00,0.00}{28.49} / \textcolor[rgb]{1.00,0.00,0.00}{0.780} & 28.18 / 0.770 & 28.19 / 0.772 & 28.15 / 0.772 & \textcolor[rgb]{0.00,0.00,1.00}{28.35} / \textcolor[rgb]{0.00,0.00,1.00}{0.777} \\\hline

    &$\times 2$ & 29.55 / 0.843 &  31.34 / 0.887&31.89 / 0.896 & --  & \textcolor[rgb]{0.00,0.00,1.00}{32.01} / \textcolor[rgb]{0.00,0.00,1.00}{0.897} &  31.80 / 0.895 & 31.90 / 0.896  &  \textcolor[rgb]{1.00,0.00,0.00}{32.05} / \textcolor[rgb]{1.00,0.00,0.00}{0.898} \\
 BSD100 &$\times 3$ & 27.20 / 0.738 &  28.40 / 0.786 &28.82 / 0.798 & -- & \textcolor[rgb]{0.00,0.00,1.00}{28.91} / \textcolor[rgb]{0.00,0.00,1.00}{0.799} & 28.82 / 0.798& 28.87 / \textcolor[rgb]{0.00,0.00,1.00}{0.799} & \textcolor[rgb]{1.00,0.00,0.00}{28.97} / \textcolor[rgb]{1.00,0.00,0.00}{0.803} \\
  &$\times 4$ & 25.96 / 0.667&  26.90 / 0.710 &27.28 / 0.726 & \textcolor[rgb]{1.00,0.00,0.00}{27.58} / \textcolor[rgb]{1.00,0.00,0.00}{0.735}  & 27.35 / 0.726 & 27.32 / 0.727 & 27.34 / 0.728 & \textcolor[rgb]{0.00,0.00,1.00}{27.49} / \textcolor[rgb]{0.00,0.00,1.00}{0.734}\\\hline

    &$\times 2$ & 26.66 / 0.841 &   29.53 / 0.897&30.76 / 0.914 & --  & \textcolor[rgb]{0.00,0.00,1.00}{31.02} / \textcolor[rgb]{0.00,0.00,1.00}{0.916} & 30.82 / 0.915 & 30.89 / \textcolor[rgb]{0.00,0.00,1.00}{0.916} & \textcolor[rgb]{1.00,0.00,0.00}{31.33} / \textcolor[rgb]{1.00,0.00,0.00}{0.920} \\
  Urban100  &$\times 3$ & 24.46 / 0.737 &  26.25 / 0.801 &27.13 / 0.828 & -- & \textcolor[rgb]{0.00,0.00,1.00}{27.38} / \textcolor[rgb]{0.00,0.00,1.00}{0.833} & 27.07 / 0.828 & 27.27 / \textcolor[rgb]{0.00,0.00,1.00}{0.833} & \textcolor[rgb]{1.00,0.00,0.00}{27.57} / \textcolor[rgb]{1.00,0.00,0.00}{0.840}\\
  &$\times 4$ & 23.14 / 0.657 &  24.52 / 0.722 &25.17 / 0.753 & -- & \textcolor[rgb]{0.00,0.00,1.00}{25.35} / 0.758 &  25.21 / 0.756 & 25.34 / \textcolor[rgb]{0.00,0.00,1.00}{0.761} & \textcolor[rgb]{1.00,0.00,0.00}{25.68} / \textcolor[rgb]{1.00,0.00,0.00}{0.773} \\\hline

\end{tabular}
\label{table1}\vspace{-0.25cm}
\end{table*}

\subsection{Experiments on Bicubic Degradation}

\begin{figure*}[!t]\vspace{-0.1cm}
\scriptsize{
\begin{center}
\subfigure[SRCNN (23.78dB)]
{\includegraphics[width=0.156\textwidth]{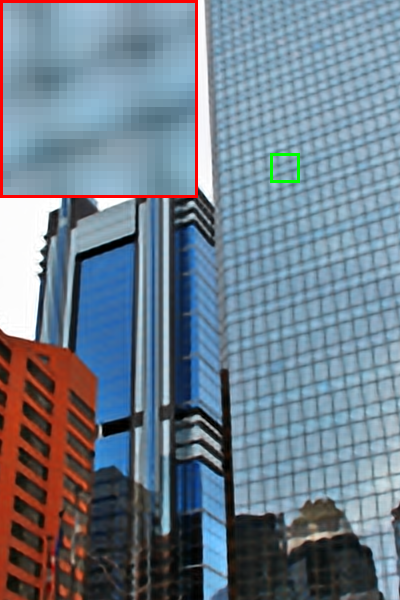}}
\hspace{0.1cm}
\subfigure[VDSR (24.20dB)]
{\includegraphics[width=0.156\textwidth]{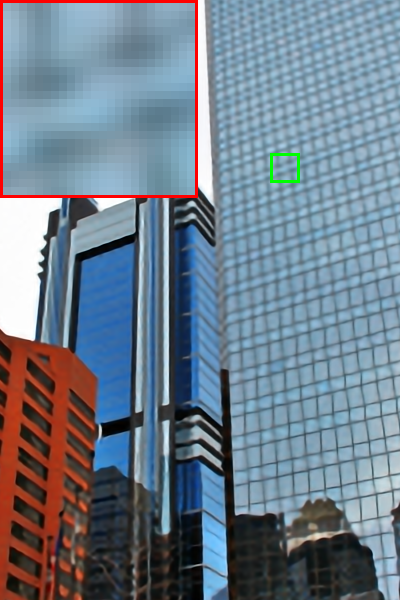}}
\hspace{0.1cm}
\subfigure[DRRN (25.11dB)]
{\includegraphics[width=0.156\textwidth]{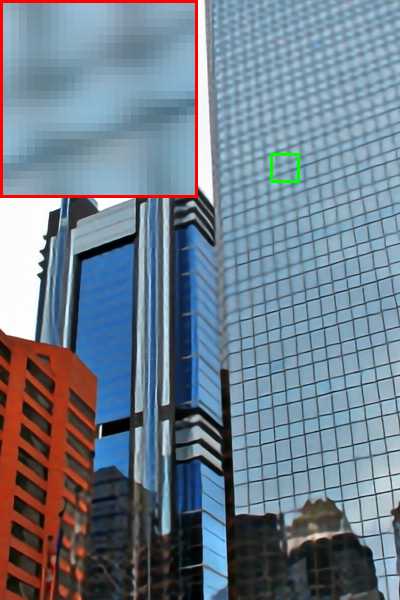}}
\hspace{0.1cm}
\subfigure[LapSR (24.47dB)]
{\includegraphics[width=0.156\textwidth]{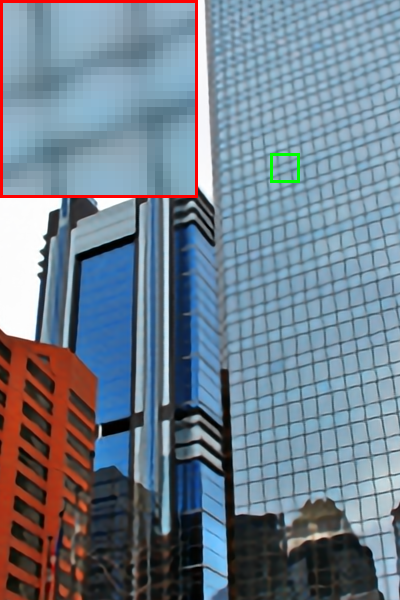}}
\hspace{0.1cm}
\subfigure[SRMD (25.09dB)]
{\includegraphics[width=0.156\textwidth]{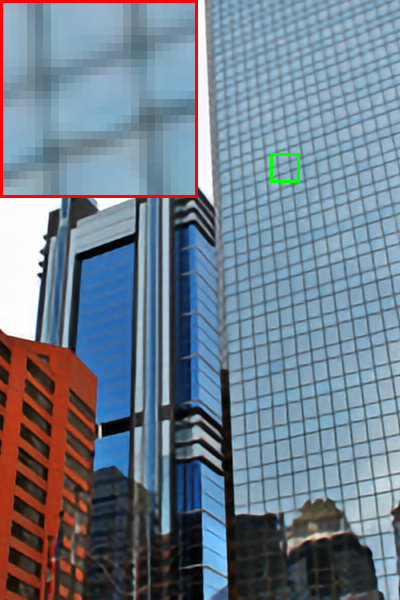}}
\hspace{0.1cm}
\subfigure[\hspace{-0.1cm}SRMDNF~(25.74dB)]
{\includegraphics[width=0.156\textwidth]{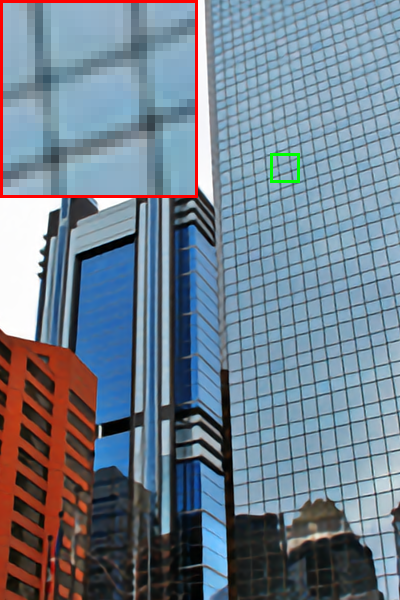}}
\caption{SISR performance comparison of different methods with scale factor 4 on image ``\emph{Img\_099}'' from Urban100.}\label{fig_bicubic1}
\end{center}}\vspace{-0.4cm}
\end{figure*}

As mentioned above, instead of handling the bicubic degradation only, our aim is to learn a single network to handle multiple degradations. However, in order to show the advantage of the dimensionality stretching strategy, the proposed method is also compared with other CNN-based methods specifically designed for bicubic degradation.

Table~\ref{table1} shows the PSNR and SSIM~\cite{wang2004image} results of state-of-the-art CNN-based SISR methods on four widely-used datasets.
As one can see, SRMD achieves comparable results with VDSR at small scale factor and outperforms VDSR at large scale factor.
In particular, SRMDNF achieves the best overall quantitative results.
Using ImageNet dataset~\cite{krizhevsky2012imagenet} to train the specific model with bicubic degradation, SRResNet performs slightly better than SRMDNF on scale factor 4.
To further compare with other methods such as VDSR, we also have trained a SRMDNF model (for scale factor 3) which operates on Y channel with 291 training images.
The learned model achieves 33.97dB, 29.96dB, 28.95dB and 27.42dB on Set5, Set14, BSD100 and Urban100, respectively. As a result,
it can still outperform other competing methods. The possible reason lies in that the SRMDNF with multiple degradations shares
the same prior in the MAP framework which facilitates the
implicit prior learning and thus benefits to PSNR improvement.
This also can explain why VDSR with multiple scales improves the performance.

For the GPU run time, SRMD spends 0.084, 0.042 and 0.027 seconds to reconstruct an HR image of size $1,024\times1,024$ for scale factors 2, 3 and 4, respectively.
As a comparison, the run time of VDSR is 0.174 second for all scale factors.
Figure~\ref{fig_bicubic1} shows the visual results of different methods.
One can see that our proposed method yields very competitive performance against other methods.

\begin{table*}[!htbp]\scriptsize
\caption{Average PSNR and SSIM results of different methods with different degradations on Set5. The best results are highlighted in \textcolor[rgb]{1.00,0.00,0.00}{red} color. The results highlighted in \textcolor[rgb]{0.45,0.45,0.45}{gray} color indicate unfair comparison due to mismatched degradation assumption.} \vspace{-0.1cm}%Red color indicates the best performance and blue color indicates the second best performance
\center
\begin{tabular}{|p{0.84cm}<{\centering}|p{0.8cm}<{\centering}|p{0.8cm}<{\centering}|c|p{1.4cm}<{\centering}| c|c|c|p{1.8cm}<{\centering}|}
  \hline
  % after \\: \hline or  \cline{col3-col4} ...\multirow{4}{2cm}{This is a demo table}
 \multicolumn{3}{|c|}{Degradation Settings} &  VDSR~\cite{kim2015accurate} & NCSR~\cite{dong2013nonlocally}  & IRCNN~\cite{zhang2017learning}  & DnCNN~\cite{zhang2017beyond}+SRMDNF & SRMD  & SRMDNF\\\cline{1-9}
   Kernel  & Down- & Noise  &  \multicolumn{6}{c|}{\multirow{2}{*}{PSNR ($\times$2/$\times$3/$\times$4)}}   \\
    Width & sampler & Level  & \multicolumn{6}{c|}{}    \\ \hline\hline
  % &  & Bicubic & A$+$ & SelfEx & SRCNN  & VDSR  & JSCNN (Ours) \\ \hline
  0.2 & Bicubic&  0 & 37.56/33.67/31.35 & --~~/\textcolor[rgb]{0.45,0.45,0.45}{23.82}/--~~  &  37.43/33.39/31.02  & -- & 37.53/33.86/31.59  & \textcolor[rgb]{1.00,0.00,0.00}{37.79}/\textcolor[rgb]{1.00,0.00,0.00}{34.12}/\textcolor[rgb]{1.00,0.00,0.00}{31.96}  \\
  0.2 & Bicubic & 15  &\textcolor[rgb]{0.45,0.45,0.45}{26.02}/\textcolor[rgb]{0.45,0.45,0.45}{25.40}/\textcolor[rgb]{0.45,0.45,0.45}{24.70} & --  & 32.60/30.08/28.35 & 32.47/30.07/28.31 & \textcolor[rgb]{1.00,0.00,0.00}{32.76}/\textcolor[rgb]{1.00,0.00,0.00}{30.43}/\textcolor[rgb]{1.00,0.00,0.00}{28.79} & --\\
  0.2 & Bicubic & 50  &\textcolor[rgb]{0.45,0.45,0.45}{16.02}/\textcolor[rgb]{0.45,0.45,0.45}{15.72}/\textcolor[rgb]{0.45,0.45,0.45}{15.46} & --  & 28.20/26.25/24.95 & 28.20/26.27/24.93 & \textcolor[rgb]{1.00,0.00,0.00}{28.51}/\textcolor[rgb]{1.00,0.00,0.00}{26.48}/\textcolor[rgb]{1.00,0.00,0.00}{25.18} & --\\
\cline{1-9}
   1.3 & Bicubic & 0 &\textcolor[rgb]{0.45,0.45,0.45}{30.57}/\textcolor[rgb]{0.45,0.45,0.45}{30.24}/\textcolor[rgb]{0.45,0.45,0.45}{29.72} & --~~/\textcolor[rgb]{0.45,0.45,0.45}{21.81}/--~~  & 36.01/33.33/31.01  &  -- & 37.04/33.77/31.56  &  \textcolor[rgb]{1.00,0.00,0.00}{37.45}/\textcolor[rgb]{1.00,0.00,0.00}{34.16}/\textcolor[rgb]{1.00,0.00,0.00}{31.99}\\
 1.3 & Bicubic & 15 &\textcolor[rgb]{0.45,0.45,0.45}{24.82}/\textcolor[rgb]{0.45,0.45,0.45}{24.70}/\textcolor[rgb]{0.45,0.45,0.45}{24.30} & --  & 29.96/28.68/27.71 & 27.68/28.78/27.71 & \textcolor[rgb]{1.00,0.00,0.00}{30.98}/\textcolor[rgb]{1.00,0.00,0.00}{29.43}/\textcolor[rgb]{1.00,0.00,0.00}{28.21} & --\\
   1.3 & Bicubic & 50 &\textcolor[rgb]{0.45,0.45,0.45}{15.89}/\textcolor[rgb]{0.45,0.45,0.45}{15.68}/\textcolor[rgb]{0.45,0.45,0.45}{15.43}& -- & 26.69/25.20/24.42 & 24.35/25.19/24.39 & \textcolor[rgb]{1.00,0.00,0.00}{27.43}/\textcolor[rgb]{1.00,0.00,0.00}{25.82}/\textcolor[rgb]{1.00,0.00,0.00}{24.77} & --\\
\cline{1-9}
2.6 & Bicubic & 0 &\textcolor[rgb]{0.45,0.45,0.45}{26.37}/\textcolor[rgb]{0.45,0.45,0.45}{26.31}/\textcolor[rgb]{0.45,0.45,0.45}{26.28} & --~~/\textcolor[rgb]{0.45,0.45,0.45}{21.46}/--~~  & 32.07/31.09/30.06  &  -- & 33.24/32.59/31.20  &  \textcolor[rgb]{1.00,0.00,0.00}{34.12}/\textcolor[rgb]{1.00,0.00,0.00}{33.02}/\textcolor[rgb]{1.00,0.00,0.00}{31.77} \\
  2.6 & Bicubic & 15  &\textcolor[rgb]{0.45,0.45,0.45}{23.09}/\textcolor[rgb]{0.45,0.45,0.45}{23.07}/\textcolor[rgb]{0.45,0.45,0.45}{22.98} & -- & 26.44/25.67/24.36 & ~~~~--~~/21.33/23.85& \textcolor[rgb]{1.00,0.00,0.00}{28.48}/\textcolor[rgb]{1.00,0.00,0.00}{27.55}/\textcolor[rgb]{1.00,0.00,0.00}{26.82} & -- \\
  2.6 & Bicubic & 50 &\textcolor[rgb]{0.45,0.45,0.45}{15.58}/\textcolor[rgb]{0.45,0.45,0.45}{15.43}/\textcolor[rgb]{0.45,0.45,0.45}{15.23} & --  & 22.98/22.16/21.43 & ~~~~--~~/19.03/21.15 & \textcolor[rgb]{1.00,0.00,0.00}{25.85}/\textcolor[rgb]{1.00,0.00,0.00}{24.75}/\textcolor[rgb]{1.00,0.00,0.00}{23.98} & --\\\hline\hline

 1.6 & Direct &  0 &--~~/\textcolor[rgb]{0.45,0.45,0.45}{30.54}/~~-- & --~~/33.02/~~--  & --~~/33.38/~~-- & -- & --~~/33.74/~~-- & --~~/\textcolor[rgb]{1.00,0.00,0.00}{34.01}/~~-- \\\hline
\end{tabular}
\label{table2}\vspace{-0.2cm}
\end{table*}

\begin{figure*}[!htbp]\vspace{-0.1cm}
\scriptsize{
\begin{center}
\subfigure[Ground-truth]
{\includegraphics[width=0.1585\textwidth]{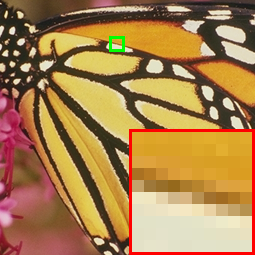}}
\hspace{0.03cm}
\subfigure[VDSR (24.73dB)]
{\includegraphics[width=0.1585\textwidth]{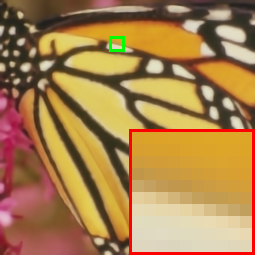}}
\hspace{0.03cm}
\subfigure[NCSR (28.01dB)]
{\includegraphics[width=0.1585\textwidth]{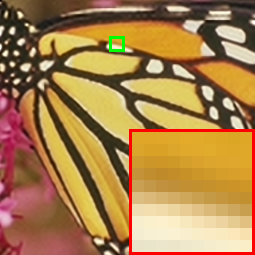}}
\hspace{0.03cm}
\subfigure[IRCNN (29.32dB)]
{\includegraphics[width=0.1585\textwidth]{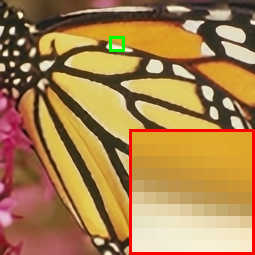}}
\hspace{0.03cm}
\subfigure[SRMD (29.79dB)]
{\includegraphics[width=0.1585\textwidth]{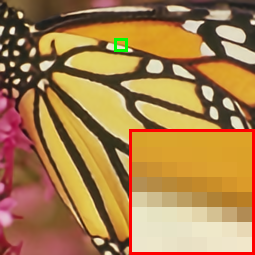}}
\hspace{0.03cm}
\subfigure[SRMDNF (30.34dB)]
{\includegraphics[width=0.1585\textwidth]{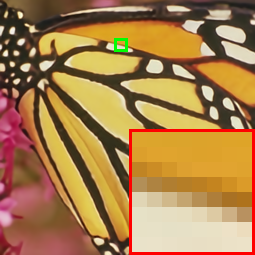}}
\caption{SISR performance comparison on image ``\emph{Butterfly}'' from Set5. The degradation involves 7$\times$7 Gaussian kernel
with width 1.6 and direct downsampler with scale factor 3. Note that the comparison with VDSR is unfair because of degradation mismatch.}\label{fig_bicubic2}
\end{center}}\vspace{-0.78cm}
\end{figure*}

%\vspace{-0.1cm}
\subsection{Experiments on General Degradations}
In this subsection, we evaluate the performance of the proposed method on general degradations.
The degradation settings are given in Table~\ref{table2}.
We only consider the isotropic Gaussian blur kernel for an easy comparison.
To further show the scalability of the proposed method, another widely-used degradation~\cite{dong2013nonlocally} which involves 7$\times$7 Gaussian kernel
with width 1.6 and direct downsampler with scale factor 3 is also included.
We compare the proposed method with VDSR, two model-based methods (\ie, NCSR~\cite{dong2013nonlocally} and IRCNN~\cite{zhang2017learning}), and a cascaded denoising-SISR method (\ie, DnCNN~\cite{zhang2017beyond}+SRMDNF).

The quantitative results of different methods with different degradations on Set5 are provided in Table~\ref{table2}, from which we have observations and analyses as follows.
First, the performance of VDSR deteriorates seriously when the assumed bicubic degradation deviates from the true one.
Second, SRMD produces much better results than NCSR and IRCNN, and outperforms DnCNN+SRMDNF. In particular, the PSNR gain of SRMD over DnCNN+SRMDNF increases with the kernel width
which verifies the advantage of joint denoising and super-resolution.
Third, by setting proper blur kernel, the proposed method delivers good performance in handling the degradation with direct downsampler. The visual comparison is given in Figure~\ref{fig_bicubic2}.
One can see that NCSR and IRCNN produce more visually pleasant results than VDSR since their assumed degradation matches the true one. However, they cannot recover edges as sharper as SRMD and SRMDNF.

\begin{figure}[!t]%\setcounter{subfigure}{-6}
\begin{center}
\subfigure[]
{\includegraphics[width=0.188\textwidth]{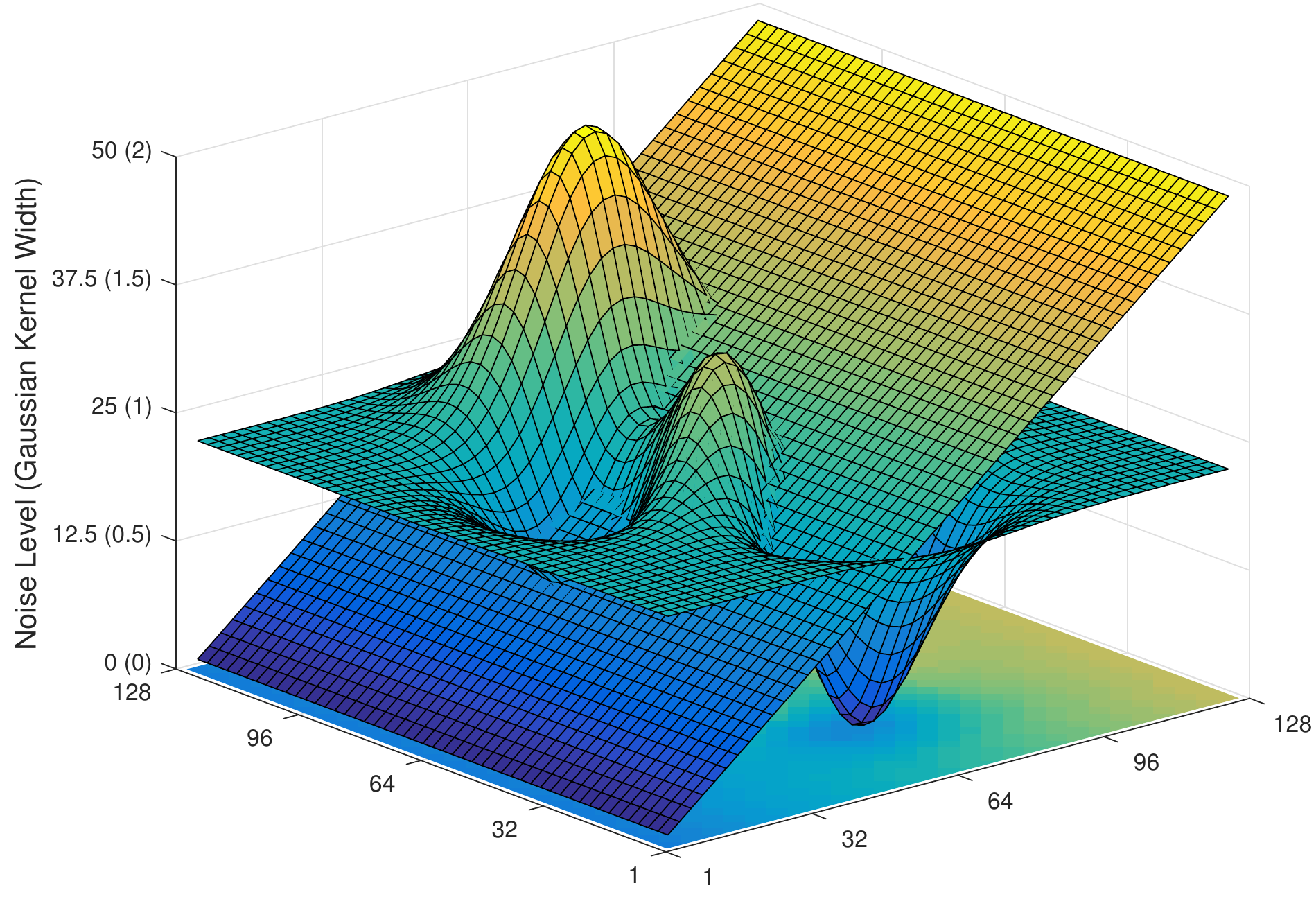}}
\subfigure[]
{\includegraphics[width=0.135\textwidth]{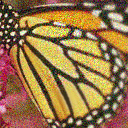}}
\hspace{0.05cm}
\subfigure[]
{\includegraphics[width=0.135\textwidth]{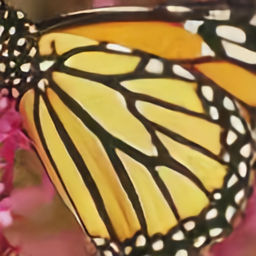}}
\caption{An example of SRMD on dealing with spatially variant degradation. (a) Noise level and Gaussian blur kernel width maps. (b) Zoomed LR image. (c) Results of SRMD with scale factor 2.}\label{fig_sv1}
\end{center}\vspace{-0.75cm}
\end{figure}

%\vspace{-0.01cm}
\subsection{Experiments on Spatially Variant Degradation}

%\vspace{-0.15cm}
To demonstrate the effectiveness of SRMD for spatially variant degradation, we synthesize an LR images with spatially variant blur kernels and noise levels.
Figure~\ref{fig_sv1} shows the visual result of the proposed SRMD for the spatially variant degradations.
One can see that the proposed SRMD is effective in recovering the latent HR image.
Note that the blur kernel is assumed to be isotropic Gaussian.

\subsection{Experiments on Real Images}

Besides the above experiments on LR images synthetically downsampled from HR images with known blur kernels and corrupted by AWGN with known noise levels, we also do experiments on real LR images to demonstrate the effectiveness of the proposed SRMD. Since there are no ground-truth HR images, we only provide the visual comparison.

As aforementioned, while we also use anisotropic Gaussian kernels in training, it is generally feasible to use isotropic Gaussian for most of the real LR images in testing. To find the degradation parameters with good visual quality, we use a grid search strategy rather than adopting any blur kernel or noise level estimation methods. Specifically, the kernel width is uniformly sampled from 0.1 to 2.4 with a stride of 0.1, and the noise level is from 0 to 75 with stride 5.

\begin{figure}[!t]%\setcounter{subfigure}{-9}
\begin{center}\vspace{0.2cm}
\subfigure[LR image]
{\includegraphics[width=0.115\textwidth]{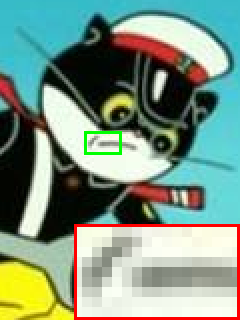}}
\subfigure[VDSR~\cite{kim2015accurate}]
{\includegraphics[width=0.115\textwidth]{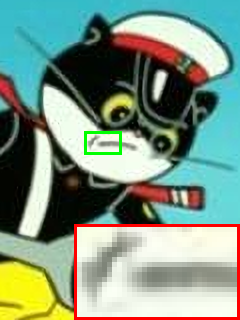}}
\subfigure[Waifu2x~\cite{waifu2x}]
{\includegraphics[width=0.115\textwidth]{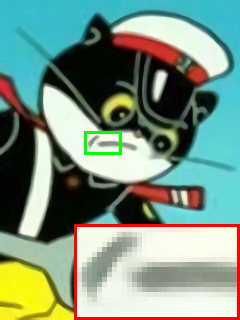}}
\subfigure[SRMD]
{\includegraphics[width=0.115\textwidth]{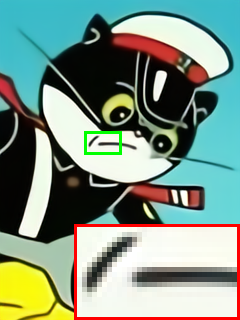}}
\caption{SISR results on image ``\emph{Cat}'' with scale factor 2.}\label{fig_real1}
\end{center}\vspace{-0.4cm}
\end{figure}

\begin{figure}[!t]
\begin{center}\vspace{-0.176cm}
\subfigure[LR image]
{\includegraphics[width=0.232\textwidth]{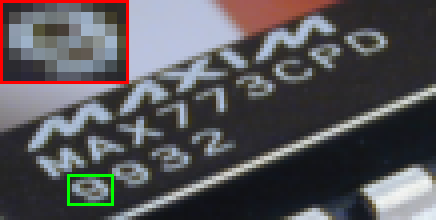}}
\hspace{0.04cm}
\subfigure[VDSR~\cite{kim2015accurate}]
{\includegraphics[width=0.232\textwidth]{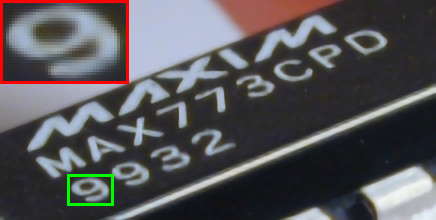}}

\vspace{-0.23cm}
\subfigure[SelfEx~\cite{huang2015single}]
{\includegraphics[width=0.232\textwidth]{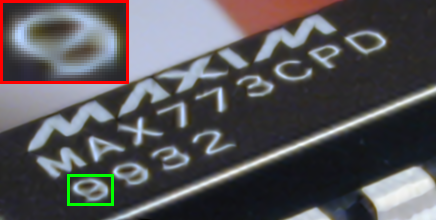}}
\hspace{0.04cm}
\subfigure[SRMD]
{\includegraphics[width=0.232\textwidth]{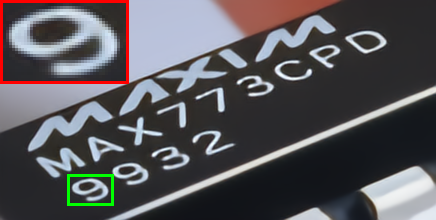}}
\caption{SISR results on real image ``\emph{Chip}'' with scale factor 4.}\label{fig_real2}
\end{center}\vspace{-0.7cm}
\end{figure}

Figures~\ref{fig_real1} and~\ref{fig_real2} illustrate the SISR results on two real LR images ``\emph{Cat}'' and  ``\emph{Chip}'', respectively.
The VDSR~\cite{kim2015accurate} is used as one of the representative CNN-based methods for comparison.
For image ``\emph{Cat}'' which is corrupted by compression artifacts, Waifu2x~\cite{waifu2x} is also used for comparison.
For image ``\emph{Chip}'' which contains repetitive structures, a self-similarity based method SelfEx~\cite{huang2015single} is also included for comparison.

It can be observed from the visual results that SRMD can produce much more visually plausible HR images than the competing methods.
Specifically, one can see from Figure~\ref{fig_real1} that the performance of VDSR is severely affected by the compression artifacts.
While Waifu2x can successfully remove the compression artifacts, it fails to recover sharp edges.
In comparison, SRMD can not only remove the unsatisfying artifacts but also produce sharp edges.
From Figure~\ref{fig_real2}, we can see that VDSR and SelfEx both tend to produce over-smoothed results,
whereas SRMD can recover sharp image with better intensity and gradient statistics of clean images~\cite{pan2014deblurring}.

\section{Conclusion}

In this paper, we proposed an effective super-resolution network with high scalability of handling multiple degradations via a single model. Different from existing CNN-based SISR methods, the proposed super-resolver takes both LR image and its degradation maps as input. Specifically, degradation maps are obtained by a simple dimensionality stretching of the degradation parameters (\ie, blur kernel and noise level). The results on synthetic LR images demonstrated that the proposed super-resolver can not only produce state-of-the-art results on bicubic degradation but also perform favorably on other degradations and even spatially variant degradations. Moreover, the results on real LR images showed that the proposed method can reconstruct visually plausible HR images. In summary, the proposed super-resolver offers a feasible solution toward practical CNN-based SISR applications.

\section{Acknowledgements}

This work is supported by National Natural Science Foundation of China (grant no.~61671182, 61471146), HK RGC General Research
Fund (PolyU 152240/15E) and PolyU-Alibaba Collaborative
Research Project ``Quality Enhancement of Surveillance Images and Videos''.
We gratefully acknowledge the support from NVIDIA Corporation for providing us the Titan Xp GPU used in this research.

\clearpage

{\small
\bibliographystyle{ieee}
\bibliography{egbib}
}

\end{document}